\PassOptionsToPackage{hyphens}{url}
\RequirePackage[svgnames,table]{xcolor}
\documentclass[11pt,letterpaper,logo]{yalearxiv}

\usepackage{natbib}
\setcitestyle{authoryear,round}

\newif\ifreplacecolor
\replacecolorfalse
\definecolor{replacepurple}{RGB}{126,0,92}
\newcommand{\est}[1]{\ifreplacecolor\textcolor{replacepurple}{#1}\else#1\fi}

\newcommand{\resultclaim}[1]{\ifreplacecolor\textcolor{replacepurple}{#1}\else#1\fi}

\newcommand{\HistoricalPapers}{380}
\newcommand{\HistoricalStates}{1,900}
\newcommand{\HistoricalIDPapers}{320}
\newcommand{\HistoricalTestPapers}{150}
\newcommand{\ConfirmationPapers}{271}

\newcommand{\HistoricalModelJudgments}{35,292}
\newcommand{\DistinctModelCalls}{77,929}
\newcommand{\InitialUnresolvedShare}{96.3\%}
\newcommand{\InitialUnresolvedMean}{16.62}
\newcommand{\InitialActiveMean}{17.27}

\newcommand{\JudgeLuna}{GPT-5.6 Luna}
\newcommand{\JudgeTerra}{GPT-5.6 Terra}
\newcommand{\JudgeSol}{GPT-5.6 Sol}

\newcommand{\PanelAdjudicationRate}{\est{30.3\%}}
\newcommand{\PanelAlphaValidity}{\est{0.74}}

\newcommand{\PanelAlphaMateriality}{\est{0.52}}
\newcommand{\DatasetListPriceEquivalent}{\est{\$9,700}}

\newcommand{\EThreeCalls}{\est{8.0}}
\newcommand{\EThreeTokens}{\est{8.9k}}

\newcommand{\EThreeOmitRate}{\est{14.4\%}}

\newcommand{\EThreeOverRate}{\est{15.5\%}}
\newcommand{\EThreeCoverage}{\est{95.4\%}}
\newcommand{\EThreeActive}{\est{17.6}}

\newcommand{\MatchedCalls}{\est{23.7}}
\newcommand{\MatchedTokens}{\est{24.9k}}

\newcommand{\MatchedOmitRate}{\est{13.7\%}}

\newcommand{\MatchedOverRate}{\est{17.3\%}}
\newcommand{\MatchedCoverage}{\est{96.1\%}}
\newcommand{\MatchedActive}{\est{21.9}}

\newcommand{\ERCalls}{\est{23.5}}
\newcommand{\ERTokens}{\est{24.7k}}

\newcommand{\EROmitRate}{\est{12.5\%}}

\newcommand{\EROverRate}{\est{8.1\%}}
\newcommand{\ERCoverage}{\est{95.1\%}}
\newcommand{\ERActive}{\est{12.9}}

\newcommand{\OverDiff}{\est{$-7.4$ percentage points}}
\newcommand{\OverCI}{\est{[$-11.6$, $-3.1$]}}

\newcommand{\CoverageDiff}{\est{$-0.3$ percentage points}}
\newcommand{\CoverageCI}{\est{[$-1.1$, $0.5$]}}
\newcommand{\ActiveDiff}{\est{$-4.7$}}

\newcommand{\StopCount}{\est{142}}
\newcommand{\StopCoverage}{\est{52.4\%}}

\newcommand{\StopFailures}{\est{8}}
\newcommand{\StopRisk}{\est{5.6\%}}
\newcommand{\StopUCB}{\est{9.9\%}}

\newcommand{\AllStopUCB}{\est{16.4\%}}

\newcommand{\MatchedCtrlRecall}{\est{96.7\%}}
\newcommand{\MatchedCtrlFP}{\est{10.8\%}}
\newcommand{\ERCtrlRecall}{\est{95.8\%}}
\newcommand{\ERCtrlFP}{\est{3.3\%}}

\newcommand{\RemainingHighShare}{\est{4.0\%}}

\newcommand{\AblFullOver}{\est{8.1}}

\newcommand{\AblNoRevisionOver}{\est{14.2}}

\newcommand{\AblNoBlindOmit}{\est{17.5}}

\newcommand{\AblNoStateOmit}{\est{16.7}}

\newcommand{\AgreementReductionShare}{\est{75.0\%}}

\newcommand{\SelSizeOnlyUCB}{\est{13.2}}
\newcommand{\SelNoSizeCoverage}{\est{50.6}}
\newcommand{\SelNoSizeUCB}{\est{9.8}}

\newcommand{\method}{\textsc{EquiReview-R}}
\newcommand{\eThree}{\textsc{E3}}
\newcommand{\eThreeMatched}{\textsc{E3-Matched}}
\newcommand{\diag}{\textsc{DIAG}}
\newcommand{\simple}{\textsc{Simple}}
\newcommand{\dataset}{\textsc{ReviewTrace}}

\title{More Criticism Does Not Make a Better Review: EquiReview-R}
\runningtitle{More Criticism Does Not Make a Better Review}
\newcommand{\emailaddr}[1]{\href{mailto:#1}{\texttt{#1}}}
\author{Zexing Zhang, Jichao Li, Tianyang Lei, Yude Fu, and Yang Kewei\\
College of Systems Engineering, National University of Defense Technology, Changsha 410073, China\\
{\small \emailaddr{zhangzexing@nudt.edu.cn}, \emailaddr{lijichao09@nudt.edu.cn}\\
\emailaddr{leitianyang20@163.com}, \emailaddr{fuyude22@nudt.edu.cn}, \emailaddr{kayyang27@nudt.edu.cn}}}

\hypersetup{
  colorlinks=true,
  linkcolor=blue!50!black,
  citecolor=blue!50!black,
  urlcolor=blue!50!black,
  pdftitle={More Criticism Does Not Make a Better Review: EquiReview-R},
  pdfauthor={Zexing Zhang, Jichao Li, Tianyang Lei, Yude Fu, Yang Kewei}
}
\begin{document}

\begin{abstract}
\vspace{-1mm}
{\centering\section*{Abstract}}
AI reviewers can now produce many specific criticisms, but more criticism is not necessarily a better review. A review may miss a consequential weakness or retain an allegation that available evidence does not support. These failures require opposite corrections, yet generation-oriented systems and aggregate measures obscure the distinction. We therefore recast AI-assisted review as evidence-guided refinement of a structured concern set, with omission and overcritique treated as separate risks. Building on this formulation, we introduce \method{}, which resolves existing concerns against localized evidence, searches for missing issues from independent and review-conditioned perspectives, and returns stop, continue, or defer. To expose the failure mode that motivates this design, we construct an evidence-linked trajectory corpus. Its retrospective analysis shows why revision must precede further search: nearly all concerns in a high-recall review lack a definitive evidential disposition, while an earlier refinement mechanism cannot revise them. \resultclaim{On a frozen cohort of previously unseen papers, \method{} satisfies the prespecified non-inferiority criterion for major omission, reduces major overcritique from \EThreeOverRate{} to \EROverRate{}, and attains a one-sided omission upper bound of \StopUCB{} while stopping on \StopCoverage{} of papers.} Computation-matched controls, controlled pairs, and ablations show that the gain comes from revision rather than extra inference or shorter output. We release the corpus as \dataset{}, an evidence-linked resource for studying review revision, disagreement, and provenance.
\end{abstract}

\maketitle

\section{Introduction}
AI assistance is already changing scientific peer review. In a large randomized study at ICLR 2025, model-generated feedback led reviewers to revise real reports and engage more deeply with author responses \citep{thakkar2026randomized}. Corpus studies likewise find that language models increasingly modify conference reviews \citep{liang2024monitoring}. A complementary study of Nature-family papers asked domain scientists to assess individual human and AI criticisms. AI reviewers surfaced issues that humans missed, but they also overlapped strongly with one another and were overly critical about minor points \citep{kim2026limits}. Together, these results suggest a shift in the central technical problem. As review agents examine more aspects of a paper and generate more candidate objections, the bottleneck moves from producing criticism to deciding which criticism remains justified after the evidence is checked.

A simple example shows why this distinction matters. Suppose one review overlooks the absence of a matched baseline. The appropriate correction is to add a concern. A second review alleges data leakage even though the paper documents a clean split. The appropriate correction is to remove or narrow the allegation. The first review lacks coverage. The second imposes an unsupported burden on authors and readers. A system can improve one error while worsening the other, even when its output becomes longer and appears more thorough.

We therefore study review improvement as the revision of a structured concern set. Each concern states one alleged scientific failure, identifies the relevant part of the paper, and records the evidence needed to resolve it. The set should expand when a material issue is missing, but it should also contract when a concern is refuted, duplicated, already answered, or broader than the evidence warrants. This view separates two paper-level risks. \emph{Major omission} captures consequential issues that are absent or remain unresolved when review ends. \emph{Major overcritique} captures consequential allegations that should be removed from or materially narrowed in the visible review.

\begin{figure*}[t]
\centering
\includegraphics[width=\textwidth]{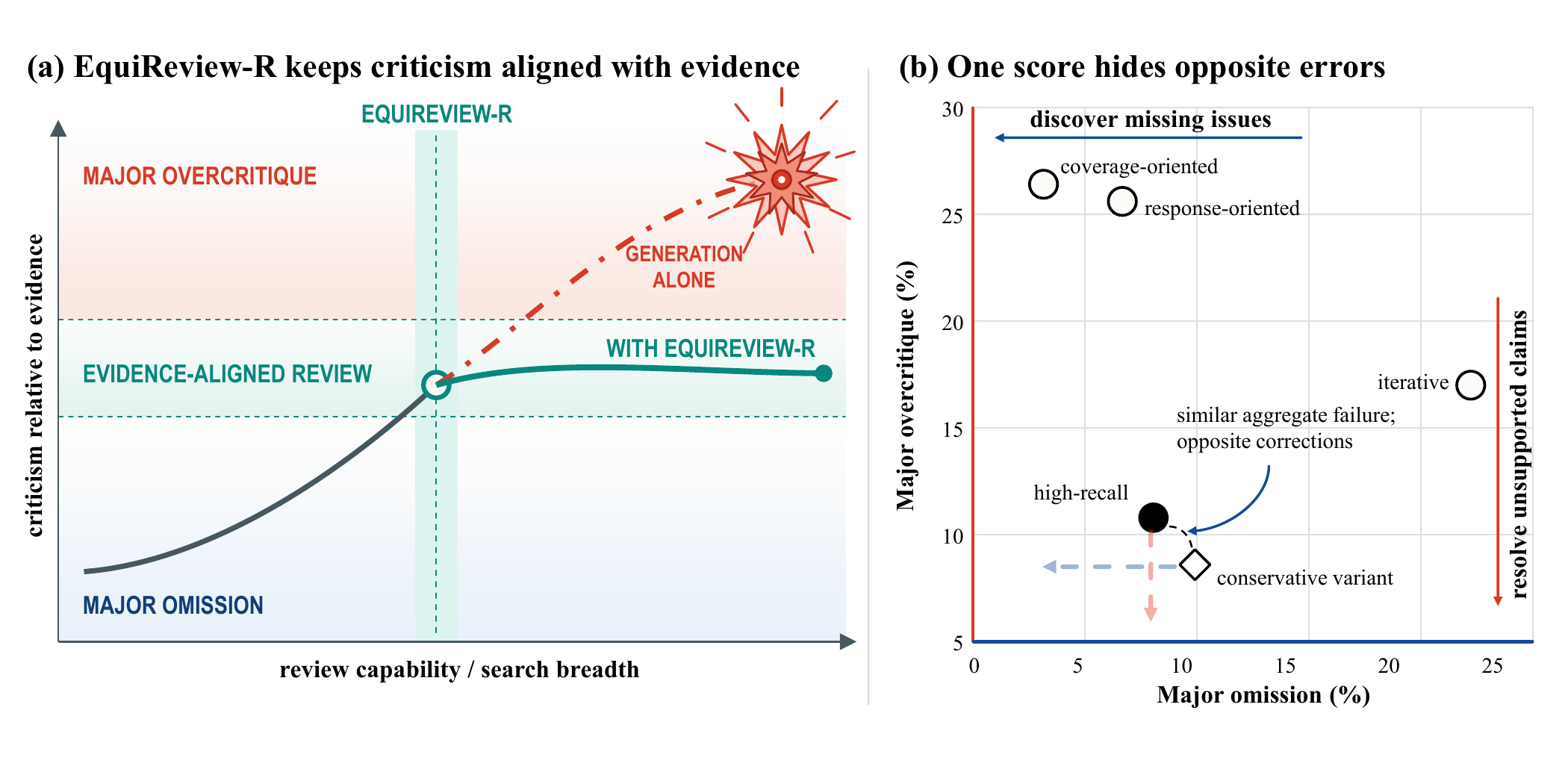}
\caption{The bottleneck shifts from finding criticism to resolving it. Additional search can expose missing issues, but it can also accumulate claims that have not survived evidential scrutiny. Reliable refinement must support both directions of correction. Retrospective systems occupy different regions of the omission-overcritique plane, so a one-dimensional failure label does not reveal which correction is needed.}
\label{fig:teaser}
\end{figure*}

To study this process directly, we constructed \dataset{} rather than extracting static reviews from an existing corpus. The resource records how each concern is proposed, challenged, related to other concerns, revised, and judged. Its retrospective trajectories expose a structural failure that aggregate scores conceal. In the high-recall review state, \InitialUnresolvedShare{} of concerns had not yet been supported, narrowed, or rejected. A later refinement mechanism left every initial concern unchanged because its revision step only revisited concerns that had already received a provisional disposition. Figure~\ref{fig:teaser} shows the broader consequence. Systems with similar aggregate failure rates can occupy very different positions in the omission-overcritique plane and therefore require different corrections.

This diagnosis leads to a direct design principle: revise the current review before expanding it. We propose \method{}, which first re-evaluates every unresolved concern against localized support, counterevidence, and an explicit resolution condition. It then freezes the revised state and searches for omissions from two complementary perspectives. One search is independent of the current review, while the other uses the revised review to target aspects it does not represent. Finally, a selective procedure returns \emph{stop}, \emph{continue}, or \emph{defer} according to unresolved consequence, evidential completeness, and judgment uncertainty. This order matters. Revision controls unsupported criticism, complementary search protects recall, and selective stopping prevents unresolved high-consequence questions from being mistaken for a finished review.

The evaluation follows the same logic. Missing concerns are sought only after a system freezes its review and stopping decision, so the answer cannot define the test. Strict coverage requires the same alleged failure and resolution condition, and every conditional omission result is paired with stopping coverage. Computation-matched controls and minimally different issue and clean-control pairs test whether gains come from revision rather than extra inference or a general preference for saying less.

Our contributions are:
\begin{itemize}
\setlength{\itemsep}{0pt}
\setlength{\parsep}{0pt}
\setlength{\topsep}{1pt}
\setlength{\parskip}{0pt}
\item We separate omission from overcritique and show that their union cannot identify the needed correction.
\item We propose \method{}, combining evidence-guided revision, complementary discovery, and selective stopping in a reconstructable state.
\item We evaluate it on a frozen cohort with matched computation, independent omission candidates, controlled pairs, and ablations.
\item We release \dataset{}, an evidence-linked corpus of concern trajectories and independent judgments.
\end{itemize}

\section{Related Work}
\paragraph{AI-assisted scientific review.}
Early resources such as PeerRead enabled review text and score prediction \citep{kang2018peerread}. Later work examined the usefulness and real-world uptake of model-generated feedback \citep{yuan2022automate,liang2024feedback,liang2024monitoring,thakkar2026randomized}. Modern systems improve generation through multi-stage analysis, retrieval, response-based verification, or proactive investigation. SEA consolidates multiple reviews \citep{yu2024sea}; DeepReview produces structured, evidence-rich reports \citep{zhu2025deepreview}; \diag{} and \eThree{} emphasize specific weaknesses and issue-level backtesting \citep{zou2026diagpaper,chaudhuri2026e3}; and ProReviewer maintains a structured log to guide active investigation \citep{fang2026proreviewer}. These approaches expand or organize the criticism a system can produce. Our object is the resulting concern set, and our question is which concerns should survive evidential scrutiny before further search or a stopping decision.

\paragraph{Evaluating review content.}
Scientific reviews have been evaluated through score agreement, review-response questions, overlap with observed feedback, concern matching, and attention across paper facets \citep{zhou2024reliable,liang2024feedback,jin2026concern,li2026beyond,shin2025blindspots}. CriticEval similarly decomposes critique quality across tasks and dimensions \citep{lan2024criticeval}. These perspectives reveal whether individual comments are correct, useful, or focused on appropriate aspects. They do not by themselves determine whether a persistent set of comments is sufficient to end review or contains claims that evidence no longer supports or that should be narrowed or merged. We make these revision actions and their paper-level consequences the primary evaluation object.

\paragraph{Revision, robustness, and selective decisions.}
Self-critique and tool-interactive critique can expose errors and improve model responses \citep{saunders2022selfcritique,gou2024critic}. In peer review, human-in-the-loop analyses and robustness studies emphasize correlated model errors, manipulation, and deployment risk \citep{drori2024human,ye2024risks,baumann2026stop,xin2026safereview}. Selective prediction provides a principled language for abstaining when uncertainty remains \citep{elYaniv2010selective,geifman2017selective}. Learn then Test and conformal risk control provide finite-sample procedures for evaluating prespecified risk constraints \citep{angelopoulos2025ltt,angelopoulos2024crc}. We connect these lines by applying selective risk control to a revised review state rather than to a scalar prediction.

\section{Method}
\subsection{Problem Formulation}
Let $x$ denote a paper and let $S_0$ be an initial structured review. We represent a review state as
\begin{equation}
S=(A,G,H),
\label{eq:state}
\end{equation}
where $A(S)$ is the set of concerns shown in the current review, $G(S)$ is a typed graph relating concerns as identical, overlapping, parent-child, or distinct, and $H(S)$ is an immutable history of evidence and revision actions. A concern $c\in A(S)$ contains an alleged failure, a paper location, supporting and countervailing evidence, a resolution condition, materiality, and a current status. Separating $A$ from $H$ allows the visible review to become more concise without erasing what was proposed or why it changed.

Validity is judged before materiality. A concern is \emph{major} only when resolving it could change the validity, scope, or evidential support of a principal claim, materially alter the interpretation of a central result, or affect the credibility of the main evaluation. Moderate concerns require local analysis or qualification, while minor concerns primarily affect presentation. The primary endpoints use only valid major concerns.

Let $Q_K$ be a fixed external search procedure with $K$ prespecified search opportunities. We define two paper-level losses. Given a stopping rule $g_\lambda$, $L_{\mathrm{miss}}(S;Q_K,g_\lambda)$ is one when $Q_K$ finds a valid major concern that is distinct from $A(S)$, or when $g_\lambda$ stops while a major concern remains unresolved. $L_{\mathrm{over}}(S)$ is one when $A(S)$ retains a major concern that independent judgment says should be removed or materially narrowed. We report omission only among papers that the rule stops,
\begin{equation}
\begin{aligned}
R_{\mathrm{miss}}(g_\lambda)
&=\mathbb{E}[L_{\mathrm{miss}}(S;Q_K,g_\lambda)\mid g_\lambda(S)=\mathrm{stop}],\\
C_{\mathrm{stop}}(g_\lambda)
&=\Pr[g_\lambda(S)=\mathrm{stop}].
\end{aligned}
\label{eq:risk}
\end{equation}
The two quantities must be interpreted together. Never stopping makes the conditional risk uninformative, whereas stopping every paper may violate the target.

Strict issue coverage is a secondary constraint. Let $\mathrm{Cov}(S)$ denote this quantity. A concern receives credit only when it matches both the alleged failure and the resolution condition of an independently constructed reference issue. Thus ``Theorem 1 lacks a proof of Lemma 2'' and ``Theorem 1 lacks a convergence-rate analysis'' are distinct even though they concern the same theorem. Overlap and parent-child relations receive partial credit only in sensitivity analyses.

The method can now be written as three coupled operators,
\begin{equation}
\begin{aligned}
S^- &= \mathcal{R}_{\phi}(x,S_0),\\
U &= \mathcal{D}_{\mathrm{ind}}(x)\cup
     \mathcal{D}_{\mathrm{cond}}(x,A(S^-)),\\
S^\star &= \Gamma_{\psi}(S^-,U),\qquad
 d=g_\lambda(f(S^\star)).
\end{aligned}
\label{eq:operators}
\end{equation}
Here $\mathcal{R}_{\phi}$ revises the existing review, the two $\mathcal{D}$ operators search for omissions from independent and review-conditioned perspectives, $\Gamma_{\psi}$ consolidates and admits candidates, and $g_\lambda$ returns $d\in\{\mathrm{stop},\mathrm{continue},\mathrm{defer}\}$. The design objective is to reduce overcritique while preserving issue-finding ability and nontrivial stopping coverage,
\begin{equation}
\begin{aligned}
\min_{\phi,\psi,\lambda}\quad
&\mathbb{E}[L_{\mathrm{over}}(S^\star)]\\
\text{subject to}\quad
&R_{\mathrm{miss}}(g_\lambda)\leq\alpha,
\quad C_{\mathrm{stop}}(g_\lambda)\geq c_0,\\
&\mathrm{Cov}(S^\star)\geq\mathrm{Cov}(S_0)-\epsilon.
\end{aligned}
\label{eq:objective}
\end{equation}
This constrained formulation makes explicit why simply shortening a review cannot solve the problem.

\subsection{EquiReview-R}
\begin{figure*}[t]
\centering
\includegraphics[width=\textwidth]{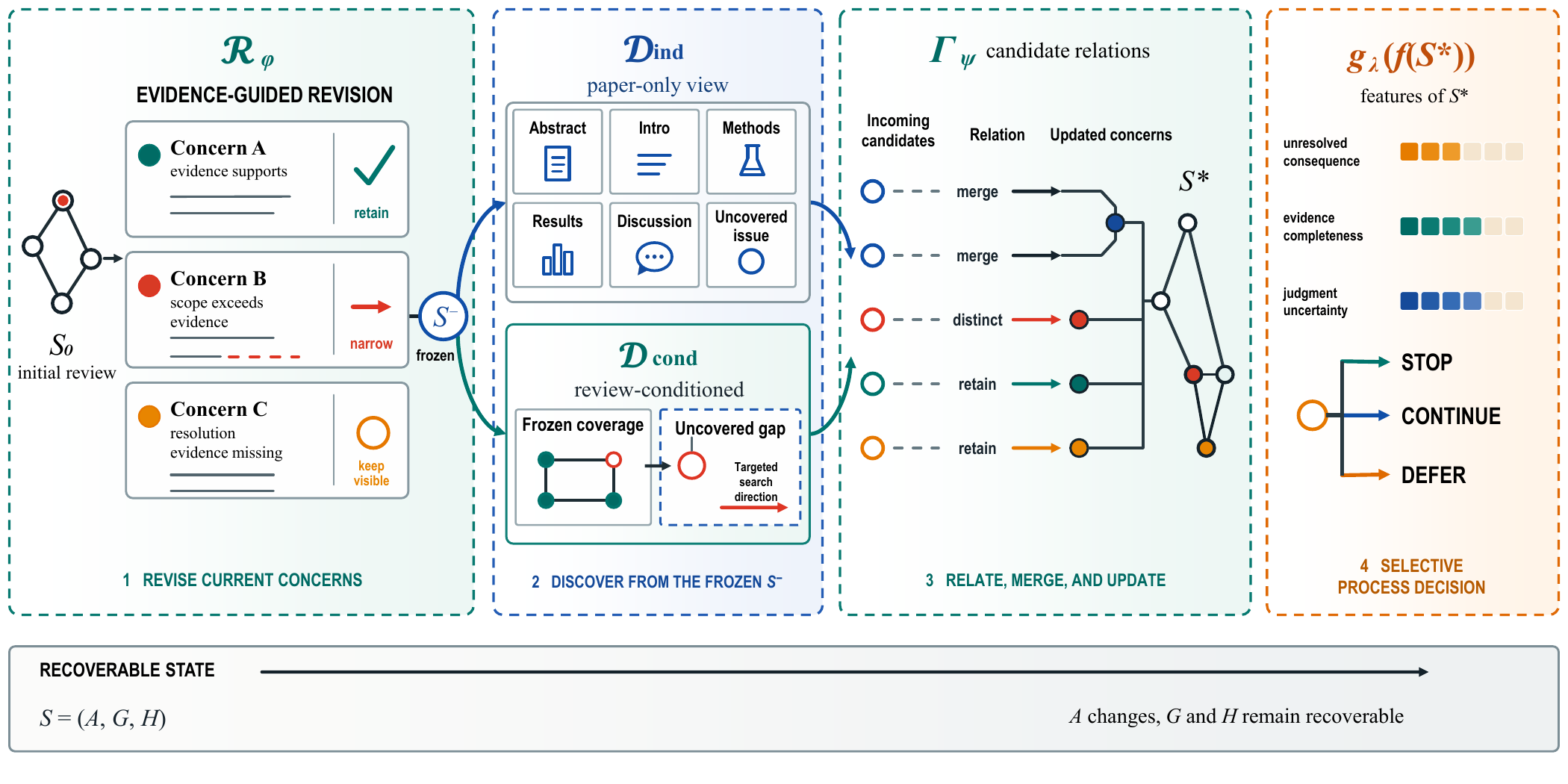}
\caption{Overview of \method{}. The revision operator $\mathcal{R}_{\phi}$ resolves the existing review before the two discovery operators search for omissions from a frozen state. $\Gamma_{\psi}$ consolidates accepted candidates, and $g_\lambda$ returns stop, continue, or defer. The reader-facing review may change, while the full trajectory remains recoverable.}
\label{fig:method}
\end{figure*}

Figure~\ref{fig:method} summarizes the four stages and their shared state. The same concern identifiers, evidence records, and relation graph connect revision, discovery, and stopping, so later stages cannot silently reinterpret what earlier stages produced.

\paragraph{Evidence-guided revision.}
For every unresolved concern, $\mathcal{R}_{\phi}$ builds an evidence record containing the alleged failure, its location, the strongest supporting evidence, the strongest counterevidence, and the condition that would settle the claim. Decomposed judgments then assign one of seven outcomes: supported, narrowed, refuted, merged, resolved, unresolved with high consequence, or unresolved with lower consequence. Supported and narrowed concerns remain visible. Refuted, merged, and resolved concerns leave the visible review but remain in $H(S)$. A high-consequence unresolved concern cannot be hidden by a presentation limit. For example, an allegation that test data influenced model selection is not accepted or removed from a split description alone. The revision record identifies where model selection is specified, whether test labels were consulted, and what evidence would settle the claim. It then retains the allegation, narrows it to a reporting ambiguity, or resolves it according to that evidence.

\paragraph{Complementary search.}
After revision, $S^-$ is frozen. $\mathcal{D}_{\mathrm{ind}}$ reads the paper without the current concern set and therefore preserves an independent route to issues the existing review may have framed away. $\mathcal{D}_{\mathrm{cond}}$ reads the same paper together with $A(S^-)$ and searches fixed scientific facets that the revised review does not yet cover. Both operate on the same frozen state. $\Gamma_{\psi}$ accepts valid material candidates, preserves graded concern relations, and consolidates duplicates before updating the visible review.

\paragraph{Selective stopping.}
The feature map $f(S^\star)$ uses only information available before external evaluation, including unresolved high-consequence concerns, evidence completeness, judgment disagreement, relation uncertainty, and recent discovery yield. The ordered rule family is fixed before confirmation labels are computed. The confirmation procedure selects the highest-coverage rule whose one-sided omission bound meets the prespecified target. At use time, that rule returns \emph{stop} only when no high-consequence concern remains unresolved. It returns \emph{continue} when another prespecified round has an actionable target, and \emph{defer} when missing artifacts or domain judgment prevent a responsible stopping decision. All three outputs return the revised review; continue and defer additionally expose the unresolved targets and required evidence.

\section{Experiments}
\subsection{Design}
We use two non-overlapping study phases. A retrospective corpus of \HistoricalPapers{} recent AI papers and \HistoricalStates{} recorded review states supports problem diagnosis, method development, and rule specification. A separate cohort of \ConfirmationPapers{} previously unseen papers is used once for confirmation after the method, search process, judgment protocol, rule ordering, and statistical analysis are frozen. The confirmation papers span machine learning, natural language processing, computer vision, and AI systems. Every system receives the same main-paper view.

The primary comparison is \eThree{}, a strong high-recall issue-level reviewer \citep{chaudhuri2026e3}. \eThreeMatched{} receives essentially the same number of effective calls and generated tokens as \method{} but repeats the generation-oriented procedure rather than revising its existing concerns. This control isolates the algorithmic contribution from added inference. \diag{} and \simple{} appear in the retrospective analysis.

\subsection{Evaluation Protocol}
External evaluation begins only after each system freezes both its review and stopping decision. An independent search constructs missing-concern candidates, and a separate review of the frozen state identifies consequential unresolved questions. The primary omission endpoint is the union of these two sources. It does not depend on the shared candidate pool. Pool coverage is secondary and is recomputed after excluding each evaluated system's own candidates. This separation prevents a system's output from defining its own success criterion.

Candidates are judged independently by \JudgeLuna{} and \JudgeTerra{}. When their categorical decisions disagree, \JudgeSol{} provides a third decision in a separate call with source information hidden. The judges do not see system identity, candidate source, or one another's rationale. Validity, identity, revision action, and materiality are elicited separately. These judgments form a repeatable model-based measurement panel rather than objective scientific truth; the analysis therefore reports disagreement, alternative reference policies, and the contribution of adjudicated cases.

The primary tests are major-omission non-inferiority, reduction in major overcritique, and the omission criterion for stopped papers. Paired binary endpoints use McNemar tests and paper-level bootstrap intervals. Candidate stopping rules are evaluated in a fixed sequence with Learn then Test family-wise control \citep{angelopoulos2025ltt}. Strict coverage, visible concern count, generated tokens, controlled pairs, component ablations, relation-policy sensitivity, and subfield heterogeneity provide secondary evidence.

\section{Results}
\subsection{Diagnosis}
The retrospective analysis first asks whether a single failure event identifies the correction a review needs. It does not. On \HistoricalTestPapers{} papers, the response-oriented reviewer and the coverage-oriented predecessor both attain low omission but high overcritique, whereas the conservative refinement variant moves in the opposite direction. The high-recall initializer is more balanced but leaves \InitialUnresolvedMean{} of \InitialActiveMean{} visible concerns per paper unresolved. Across \HistoricalIDPapers{} papers, the earlier revision mechanism changes none of these initial concerns. The dominant uncertainty is therefore not a peripheral implementation detail. It is precisely the part of the review that must be revised before additional search can be interpreted.

\subsection{Main Results}
\begin{table}[t]
\centering
\small
\setlength{\tabcolsep}{2.6pt}
\begin{tabular}{lrrrrrr}
\toprule
Method & Calls & Tok. & Omit. & Over. & Cov. & Visible \\
& /paper & /paper & $\downarrow$ & $\downarrow$ & $\uparrow$ & $\downarrow$ \\
\midrule
\eThree{} & \EThreeCalls{} & \EThreeTokens{} & \EThreeOmitRate{} & \EThreeOverRate{} & \EThreeCoverage{} & \EThreeActive{} \\
\eThreeMatched{} & \MatchedCalls{} & \MatchedTokens{} & \MatchedOmitRate{} & \MatchedOverRate{} & \textbf{\MatchedCoverage{}} & \MatchedActive{} \\
\method{} & \ERCalls{} & \ERTokens{} & \textbf{\EROmitRate{}} & \textbf{\EROverRate{}} & \ERCoverage{} & \textbf{\ERActive{}} \\
\bottomrule
\end{tabular}
\caption{Confirmation results on \ConfirmationPapers{} unseen papers. Omit. and Over. are paper-level major omission and overcritique. Cov. is strict shared-pool issue coverage. Tok. reports generated output tokens. Best outcome values are bold.}
\label{tab:main}
\end{table}

Table~\ref{tab:main} establishes the main result. \resultclaim{Relative to the high-recall initializer, \method{} meets the prespecified omission non-inferiority criterion and reduces major overcritique by \OverDiff{} (95\% CI \OverCI{}). Strict issue coverage changes by only \CoverageDiff{} (95\% CI \CoverageCI{}), while the visible review contains \ActiveDiff{} fewer concerns per paper.} The result is therefore a revision of the error profile rather than an exchange of recall for concision.

The computation-matched control provides the key counterfactual. \resultclaim{Although it receives essentially the same number of calls and output tokens, \eThreeMatched{} retains a substantially larger visible review and more than twice the major-overcritique rate of \method{}.} Additional inference is useful only when it changes the review state rather than merely extending the review.

\subsection{Efficiency}
\begin{figure}[t]
\centering
\includegraphics[width=0.96\columnwidth]{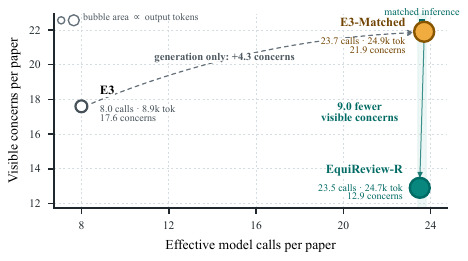}
\caption{Resource use and visible review size. Bubble area is proportional to generated output tokens. \resultclaim{At closely matched inference, \method{} returns a smaller visible review than the generation-only control.}}
\label{fig:efficiency}
\end{figure}

Figure~\ref{fig:efficiency} makes this distinction visible. \resultclaim{Generation-only inference moves the matched control upward by expanding the visible review, whereas \method{} uses comparable inference to resolve and consolidate existing claims.} The result isolates how computation is used, not merely how much is supplied.

\subsection{Selective Stopping}
\begin{figure}[t]
\centering
\includegraphics[width=\columnwidth]{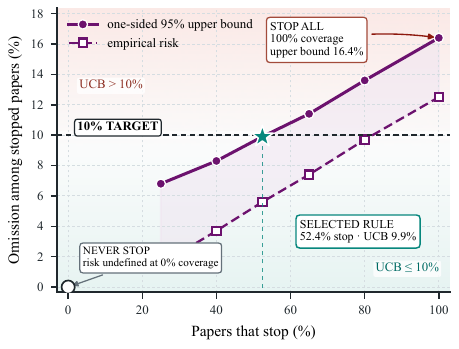}
\caption{Risk and stopping coverage for the frozen rule family. The stop-all endpoint fails the target, while never stopping provides no useful stopping decision. \resultclaim{The annotated point is the highest-coverage rule whose one-sided upper bound satisfies the 10\% omission criterion.}}
\label{fig:risk}
\end{figure}

\resultclaim{The selected rule stops on \StopCount{} of \ConfirmationPapers{} papers. It records \StopFailures{} omissions in that subset, with empirical risk \StopRisk{} and a one-sided upper bound of \StopUCB{} at \StopCoverage{} coverage. Stopping every paper yields an upper bound of \AllStopUCB{}.} Figure~\ref{fig:risk} therefore shows a nontrivial operating point rather than reliability obtained by stopping every paper or deferring nearly all of them.

The remaining papers still receive usable reviews. Continue identifies the unresolved target and evidence needed for another prespecified round, while defer exposes the high-consequence uncertainty, missing artifact, or expertise that requires human attention. Stopping is selective, but the review itself is not withheld.

\subsection{Mechanism}
\begin{figure}[t]
\centering
\includegraphics[width=\columnwidth]{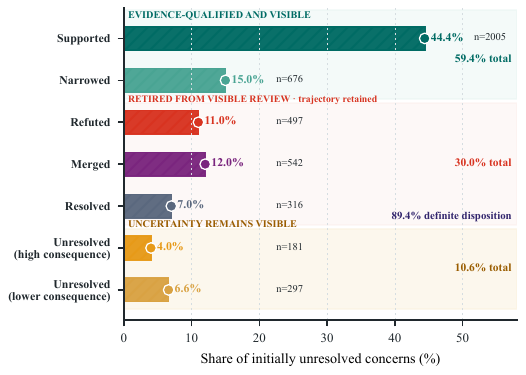}
\caption{Evidence-guided outcomes for initially unresolved concerns. \resultclaim{Most concerns receive a definite revision, while consequential uncertainty remains visible.}}
\label{fig:transitions}
\end{figure}

Figure~\ref{fig:transitions} shows how initially unresolved concerns change after evidence is considered. \resultclaim{Most receive a definite disposition, and only \RemainingHighShare{} remain unresolved with high consequence.} Supported and narrowed concerns remain visible at the scope justified by the evidence, while refuted, merged, and resolved concerns leave the visible review but remain in the trajectory. The review becomes shorter through explicit actions rather than silent deletion.

Controlled pairs provide an independent test of that interpretation. \resultclaim{\method{} recalls \ERCtrlRecall{} of inserted issues and introduces false concerns in \ERCtrlFP{} of clean controls. \eThreeMatched{} attains \MatchedCtrlRecall{} recall but raises the clean false-positive rate to \MatchedCtrlFP{}.} Thus the smaller review retains high issue sensitivity because unsupported content is corrected rather than criticism being suppressed.

\begin{figure*}[t]
\centering
\includegraphics[width=\textwidth]{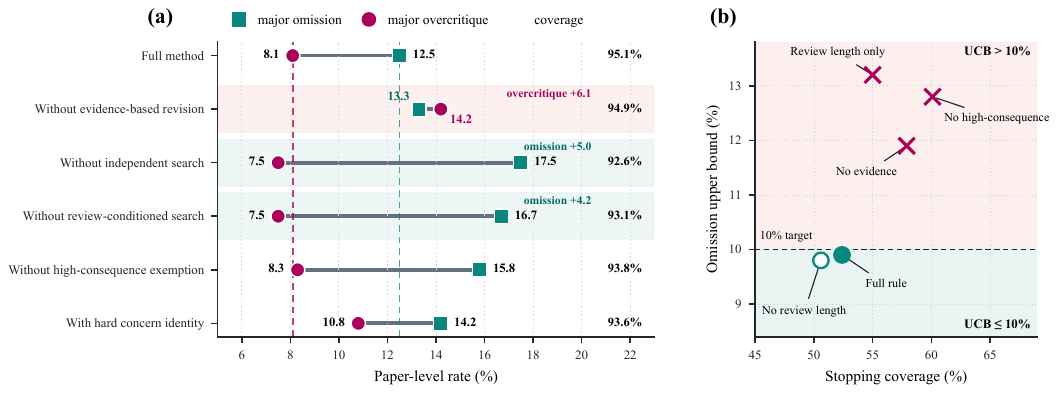}
\caption{Component and stopping-rule ablations. Left: revision primarily controls overcritique, whereas both discovery perspectives protect omission. Right: review length alone does not identify a reliable stopping point. Points below the horizontal line satisfy the omission criterion.}
\label{fig:ablation}
\end{figure*}

The ablations connect each component to the risk it is intended to control. \resultclaim{Without evidence-guided revision, overcritique rises from \AblFullOver\% to \AblNoRevisionOver\%. Removing the independent or review-conditioned search raises omission to \AblNoBlindOmit\% and \AblNoStateOmit\%, respectively. Removing the high-consequence exemption or replacing graded relations with hard identity also worsens the error profile.} The stopping-rule analysis reaches the same conclusion. \resultclaim{A rule based only on review length has a \SelSizeOnlyUCB\% omission upper bound and fails the target, whereas a rule that omits length but retains evidence and uncertainty signals attains a \SelNoSizeUCB\% bound at \SelNoSizeCoverage\% coverage.}

\subsection{Robustness}
\begin{figure}[t]
\centering
\includegraphics[width=0.94\columnwidth]{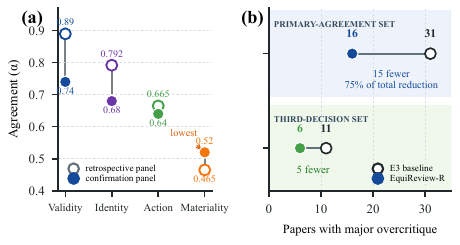}
\caption{Judgment stability and result decomposition. Agreement is lowest for materiality. \resultclaim{Most of the overcritique reduction comes from items on which the two primary judges agree.}}
\label{fig:agreement}
\end{figure}

\resultclaim{Agreement is \PanelAlphaValidity{} for validity and \PanelAlphaMateriality{} for materiality, and \PanelAdjudicationRate{} of items receive a third decision. Of the gross reduction in overcritique cases, \AgreementReductionShare{} comes from items on which the two primary judges agree about materiality. Either-judge and both-judge policies preserve the direction of the comparison.} Figure~\ref{fig:agreement} shows that materiality is the main source of measurement uncertainty, while the observed reduction is not confined to adjudicated cases.

The conclusion is also stable to alternative issue-identity policies. \resultclaim{Partial credit for overlap and parent-child relations, as well as leave-one-system-out reference pools, preserves the coverage ordering.} Thus neither looser semantic matching nor self-contributed reference wording explains the result. Subfield analyses preserve the direction within the sampled AI population, although systems papers defer more often.

\section{ReviewTrace}
We built \dataset{} as part of the study rather than deriving labels from an existing review corpus. It records each concern from first appearance through support, narrowing, merging, resolution, or removal, links every change to localized evidence, and preserves two independent judgments together with disagreement and adjudication. Existing resources primarily organize papers, static reviews, scores, meta-reviews, or sentence-level quality labels \citep{kang2018peerread,dycke2023nlpeer,shen2022mred,purkayastha2025lazyreview}. \dataset{} instead exposes the revision trajectory needed to train and evaluate systems that must correct criticism rather than only generate it.

\begin{table}[t]
\centering
\scriptsize
\setlength{\tabcolsep}{2.4pt}
\begin{tabular}{lcccc}
\toprule
Resource & Primary unit & Traj. & Rel. & Indep. \\
\midrule
PeerRead & paper/review & -- & -- & -- \\
NLPeer & paper/review version & -- & -- & -- \\
MReD & meta-review sentence & -- & -- & -- \\
LazyReview & review segment & -- & -- & -- \\
\textbf{\dataset{}} & concern trajectory & \textbf{yes} & \textbf{yes} & \textbf{yes} \\
\bottomrule
\end{tabular}
\caption{Focused structural comparison with representative peer-review resources. Traj. denotes concern-level revision trajectories, Rel. typed relations among concerns, and Indep. separately retained judgments. A dash means that the feature is not a primary released unit, not that the resource lacks value for its original task.}
\label{tab:data}
\end{table}

The release contains \HistoricalStates{} structured states and \HistoricalModelJudgments{} recorded judgments, together with normalized data specifications, evaluation code, artifact hashes, and retrieval and verification utilities. The frozen construction record contains \DistinctModelCalls{} distinct model calls. \resultclaim{Applying the public list-price schedule to that record gives a public-price equivalent of \DatasetListPriceEquivalent{} \citep{openai2026pricing}.} This investment produces more than a static collection of reviews. Each transition supplies a paired example of what changed, which evidence supported the change, and which alternative judgments remained plausible. The resource can therefore support revision-policy learning, relation-aware consolidation, disagreement modeling, provenance auditing, and selective stopping. Because the independently produced judgments are retained before adjudication, future work can study where apparent label certainty reflects consensus and where it reflects a decision policy.

\section{Discussion and Limitations}
The empirical claims are conditional on recent AI papers, the main-paper view, the fixed external search process, and the stated judgment policy. Different disciplines, supplementary artifacts, or model families can change both the concern distribution and the attainable stopping coverage. The confirmation results therefore establish reliability under a specified evaluation condition, not universal completeness.

Practical use also requires explicit reporting of computation. The matched control shows that additional generation enlarges rather than improves the visible review. At comparable inference, \method{} uses its calls to revise existing concerns and returns a smaller review. Deployment should therefore choose an operating point by both stopping coverage and the evidence still required for non-stopped papers.

Luna, Terra, and Sol are separate calls from one model family, so their errors may be correlated. Materiality also admits legitimate disagreement, as human review decisions do \citep{beygelzimer2023consistency}. Blinding system identity, decomposed labels, agreement-subset analyses, and alternative reference policies reduce avoidable circularity but do not create noise-free truth. Cross-family and domain-expert validation remain necessary before deployment. Stop, continue, and defer describe the review process rather than publication merit, and deferral must not become a rejection proxy. In use, continue should present an explicit agenda for further examination, while defer should expose the missing artifact or expertise rather than return an opaque warning. An author response can then enter as new evidence without erasing the original concern or its revision history.

\section{Conclusion}
As AI reviewers become more capable, the challenge is not only to find more weaknesses, but to determine which criticisms survive evidence, what remains missing, and when uncertainty should prevent stopping. \method{} connects evidence-guided revision, complementary discovery, and selective risk control in a reconstructable state. The results show that a review can become more concise and better supported without surrendering material issue coverage. This foundation could support auditable workflows in which model and human judgments, author responses, and editorial assessment accumulate as evidence rather than replace one another.

\clearpage
\bibliographystyle{plainnat}
\bibliography{references}

@article{angelopoulos2025ltt,
  title={Learn Then Test: Calibrating Predictive Algorithms to Achieve Risk Control},
  author={Angelopoulos, Anastasios N. and Bates, Stephen and Cand\`es, Emmanuel J. and Jordan, Michael I. and Lei, Lihua},
  journal={The Annals of Applied Statistics},
  volume={19},
  number={2},
  pages={1641--1662},
  year={2025},
  doi={10.1214/24-AOAS1998}
}

@inproceedings{angelopoulos2024crc,
  title={Conformal Risk Control},
  author={Angelopoulos, Anastasios N. and Bates, Stephen and Fisch, Adam and Lei, Lihua and Schuster, Tal},
  booktitle={International Conference on Learning Representations},
  year={2024}
}

@article{elYaniv2010selective,
  title={On the Foundations of Noise-Free Selective Classification},
  author={El-Yaniv, Ran and Wiener, Yair},
  journal={Journal of Machine Learning Research},
  volume={11},
  number={53},
  pages={1605--1641},
  year={2010}
}

@inproceedings{geifman2017selective,
  title={Selective Classification for Deep Neural Networks},
  author={Geifman, Yonatan and El-Yaniv, Ran},
  booktitle={Advances in Neural Information Processing Systems},
  volume={30},
  year={2017}
}

@article{chaudhuri2026e3,
  title={E3: Issue-Level Backtesting for Automated Research Critique},
  author={Chaudhuri, Yashwardhan and Jain, Sanyam and Mundra, Paridhi},
  journal={arXiv preprint arXiv:2605.27072},
  year={2026}
}

@article{zou2026diagpaper,
  title={DIAGPaper: Diagnosing Valid and Specific Weaknesses in Scientific Papers via Multi-Agent Reasoning},
  author={Zou, Zhuoyang and Ansari, Abolfazl and Zhang, Delvin Ce and Lee, Dongwon and Yin, Wenpeng},
  journal={arXiv preprint arXiv:2601.07611},
  year={2026}
}

@article{jin2026concern,
  title={What Makes a Good AI Review? Concern-Level Diagnostics for AI Peer Review},
  author={Jin, Ming},
  journal={arXiv preprint arXiv:2604.19998},
  year={2026}
}

@article{li2026beyond,
  title={Beyond Rating: A Comprehensive Evaluation and Benchmark for AI Reviews},
  author={Li, Bowen and Ma, Haochen and Wang, Yuxin and Yang, Jie and Chen, Xinchi and Huang, Xuanjing and Zheng, Yining and Qiu, Xipeng},
  journal={arXiv preprint arXiv:2604.19502},
  year={2026}
}

@article{kim2026limits,
  title={On the Limits and Opportunities of AI Reviewers: Reviewing the Reviews of Nature-Family Papers with 45 Expert Scientists},
  author={Kim, Seungone and others},
  journal={arXiv preprint arXiv:2605.20668},
  year={2026}
}

@article{baumann2026stop,
  title={Stop Automating Peer Review Without Rigorous Evaluation},
  author={Baumann, Joachim and Pei, Jiaxin and Koyejo, Sanmi and Hovy, Dirk},
  journal={arXiv preprint arXiv:2605.03202},
  year={2026},
  note={ICML 2026 Position Paper}
}

@inproceedings{kang2018peerread,
  title={A Dataset of Peer Reviews (PeerRead): Collection, Insights and NLP Applications},
  author={Kang, Dongyeop and Ammar, Waleed and Dalvi, Bhavana and van Zuylen, Madeleine and Kohlmeier, Sebastian and Hovy, Eduard and Schwartz, Roy},
  booktitle={Proceedings of NAACL-HLT},
  pages={1647--1661},
  year={2018},
  doi={10.18653/v1/N18-1149}
}

@article{beygelzimer2023consistency,
  title={Has the Machine Learning Review Process Become More Arbitrary as the Field Has Grown? The NeurIPS 2021 Consistency Experiment},
  author={Beygelzimer, Alina and Dauphin, Yann and Liang, Percy and Vaughan, Jennifer Wortman},
  journal={arXiv preprint arXiv:2306.03262},
  year={2023}
}

@misc{openai2026pricing,
  author={{OpenAI}},
  title={OpenAI API Pricing},
  year={2026},
  howpublished={OpenAI API documentation},
  url={https://openai.com/api/pricing/},
  note={Accessed July 22, 2026}
}

@article{yuan2022automate,
  title={Can We Automate Scientific Reviewing?},
  author={Yuan, Weizhe and Liu, Pengfei and Neubig, Graham},
  journal={Journal of Artificial Intelligence Research},
  volume={75},
  pages={171--212},
  year={2022}
}

@article{liang2024feedback,
  title={Can Large Language Models Provide Useful Feedback on Research Papers? A Large-Scale Empirical Analysis},
  author={Liang, Weixin and Zhang, Yuhui and Cao, Hancheng and Wang, Binglu and Ding, Daisy Yi and Yang, Xinyu and Vodrahalli, Kailas and He, Siyu and Smith, Daniel S. and Yin, Yian and McFarland, Daniel A. and Zou, James},
  journal={NEJM AI},
  volume={1},
  number={8},
  year={2024}
}

@inproceedings{zhou2024reliable,
  title={Is LLM a Reliable Reviewer? A Comprehensive Evaluation of LLM on Automatic Paper Reviewing Tasks},
  author={Zhou, Rui and Chen, Lin and Yu, Kai},
  booktitle={Proceedings of the 2024 Joint International Conference on Computational Linguistics, Language Resources and Evaluation (LREC-COLING 2024)},
  pages={9340--9351},
  publisher={ELRA and ICCL},
  year={2024}
}

@inproceedings{yu2024sea,
  title={Automated Peer Reviewing in Paper SEA: Standardization, Evaluation, and Analysis},
  author={Yu, Jialong and Ding, Zhicheng and Tan, Jian and Luo, Kai and Weng, Zhen and Gong, Cheng and Zeng, Lijun and Cui, Renjie and Han, Chao and Sun, Qiang and Wu, Zhiyong and Lan, Yanyan and Li, Xian},
  booktitle={Findings of the Association for Computational Linguistics: EMNLP 2024},
  pages={10164--10184},
  publisher={Association for Computational Linguistics},
  year={2024}
}

@inproceedings{zhu2025deepreview,
  title={DeepReview: Improving LLM-based Paper Review with Human-like Deep Thinking Process},
  author={Zhu, Ming and Weng, Yixuan and Yang, Linyi and Zhang, Yue},
  booktitle={Proceedings of the 63rd Annual Meeting of the Association for Computational Linguistics (Volume 1: Long Papers)},
  pages={29330--29355},
  publisher={Association for Computational Linguistics},
  year={2025}
}

@article{ye2024risks,
  title={Are We There Yet? Revealing the Risks of Utilizing Large Language Models in Scholarly Peer Review},
  author={Ye, Rui and Pang, Xiang and Chai, Jing and Chen, Jia and Yin, Zhen and Xiang, Zhen and Dong, Xin and Shao, Jun and Chen, Shuo},
  journal={arXiv preprint arXiv:2412.01708},
  year={2024}
}

@article{thakkar2026randomized,
  title={A Large-Scale Randomized Study of Large Language Model Feedback in Peer Review},
  author={Thakkar, Nitya and Yuksekgonul, Mert and Silberg, Jake and Garg, Animesh and Peng, Nanyun and Sha, Fei and Yu, Rose and Vondrick, Carl and Zou, James},
  journal={Nature Machine Intelligence},
  volume={8},
  pages={326--336},
  year={2026},
  doi={10.1038/s42256-026-01188-x}
}

@article{drori2024human,
  title={Human-in-the-Loop AI Reviewing: Feasibility, Opportunities, and Risks},
  author={Drori, Iddo and Te'eni, Dov},
  journal={Journal of the Association for Information Systems},
  volume={25},
  number={1},
  pages={98--109},
  year={2024},
  doi={10.17705/1jais.00867}
}

@article{saunders2022selfcritique,
  title={Self-Critiquing Models for Assisting Human Evaluators},
  author={Saunders, William and Yeh, Catherine and Wu, Jeff and Bills, Steven and Ouyang, Long and Ward, Jonathan and Leike, Jan},
  journal={arXiv preprint arXiv:2206.05802},
  year={2022}
}

@inproceedings{gou2024critic,
  title={CRITIC: Large Language Models Can Self-Correct with Tool-Interactive Critiquing},
  author={Gou, Zhibin and Shao, Zhihong and Gong, Yeyun and Shen, Yelong and Yang, Yujiu and Duan, Nan and Chen, Weizhu},
  booktitle={International Conference on Learning Representations},
  year={2024}
}

@inproceedings{lan2024criticeval,
  title={CriticEval: Evaluating Large-Scale Language Model as Critic},
  author={Lan, Tian and Zhang, Wenwei and Xu, Chen and Huang, Heyan and Lin, Dahua and Chen, Kai and Mao, Xian-Ling},
  booktitle={Advances in Neural Information Processing Systems},
  volume={37},
  pages={66907--66960},
  year={2024}
}

@inproceedings{liang2024monitoring,
  title={Monitoring AI-Modified Content at Scale: A Case Study on the Impact of ChatGPT on AI Conference Peer Reviews},
  author={Liang, Weixin and Izzo, Zachary and Zhang, Yaohui and Lepp, Haley and Cao, Hancheng and Zhao, Xuandong and Chen, Lingjiao and Ye, Haotian and Liu, Sheng and Huang, Zhi and McFarland, Daniel and Zou, James Y.},
  booktitle={Proceedings of the 41st International Conference on Machine Learning},
  series={Proceedings of Machine Learning Research},
  volume={235},
  pages={29575--29620},
  publisher={PMLR},
  year={2024}
}

@article{fang2026proreviewer,
  title={From Passive Generation to Investigation: A Proactive Scientific Peer Review Agent},
  author={Fang, Haishuo and Feng, Yue and Gurevych, Iryna},
  journal={arXiv preprint arXiv:2606.13349},
  year={2026}
}

@inproceedings{shin2025blindspots,
  title={Mind the Blind Spots: A Focus-Level Evaluation Framework for LLM Reviews},
  author={Shin, Hyungyu and Tang, Jingyu and Lee, Yoonjoo and Kim, Nayoung and Lim, Hyunseung and Cho, Ji Yong and Hong, Hwajung and Lee, Moontae and Kim, Juho},
  booktitle={Proceedings of the 2025 Conference on Empirical Methods in Natural Language Processing},
  pages={35630--35656},
  address={Suzhou, China},
  publisher={Association for Computational Linguistics},
  year={2025},
  doi={10.18653/v1/2025.emnlp-main.1805}
}

@article{xin2026safereview,
  title={SafeReview: Defending LLM-based Review Systems Against Adversarial Hidden Prompts},
  author={Xin, Yuan and Weng, Yixuan and Zhu, Minjun and Ling, Ying and Qin, Chengwei and Hahn, Michael and Backes, Michael and Zhang, Yue and Yang, Linyi},
  journal={arXiv preprint arXiv:2604.26506},
  year={2026}
}

@inproceedings{dycke2023nlpeer,
  title={NLPeer: A Unified Resource for the Computational Study of Peer Review},
  author={Dycke, Nils and Kuznetsov, Ilia and Gurevych, Iryna},
  booktitle={Proceedings of the 61st Annual Meeting of the Association for Computational Linguistics (Volume 1: Long Papers)},
  pages={5049--5073},
  address={Toronto, Canada},
  publisher={Association for Computational Linguistics},
  year={2023},
  doi={10.18653/v1/2023.acl-long.277}
}

@inproceedings{shen2022mred,
  title={MReD: A Meta-Review Dataset for Structure-Controllable Text Generation},
  author={Shen, Chenhui and Cheng, Liying and Zhou, Ran and Bing, Lidong and You, Yang and Si, Luo},
  booktitle={Findings of the Association for Computational Linguistics: ACL 2022},
  pages={2521--2535},
  address={Dublin, Ireland},
  publisher={Association for Computational Linguistics},
  year={2022},
  doi={10.18653/v1/2022.findings-acl.198}
}

@inproceedings{purkayastha2025lazyreview,
  title={LazyReview: A Dataset for Uncovering Lazy Thinking in NLP Peer Reviews},
  author={Purkayastha, Sukannya and Li, Zhuang and Lauscher, Anne and Qu, Lizhen and Gurevych, Iryna},
  booktitle={Proceedings of the 63rd Annual Meeting of the Association for Computational Linguistics (Volume 1: Long Papers)},
  pages={3280--3308},
  address={Vienna, Austria},
  publisher={Association for Computational Linguistics},
  year={2025},
  doi={10.18653/v1/2025.acl-long.165}
}
\end{document}